%% file: main.tex
\documentclass[oneside,11pt]{article}
\usepackage{jair, theapa, rawfonts}

\ShortHeadings{Grounding Language for Transfer in Deep Reinforcement Learning}
{Narasimhan, Barzilay \& Jaakkola}

\usepackage{times}
\usepackage{latexsym}
\usepackage{amsmath}
\usepackage{multirow}
\usepackage{url}
\DeclareMathOperator*{\argmax}{arg\,max}
\usepackage{amssymb}
\usepackage{graphicx}
\usepackage{algorithm}
\usepackage{booktabs}
\usepackage{color}
\usepackage{subcaption}
\usepackage{tablefootnote}
\usepackage{breqn}
\usepackage[noend]{algpseudocode}
\usepackage{url}
\usepackage{arydshln}
\usepackage{dblfloatfix}
\usepackage[framemethod=tikz]{mdframed}
\usepackage{enumitem}
\usepackage{dashrule}

\definecolor{boxcolor}{rgb}{0,0,0} 
\newmdenv[innerlinewidth=0.5pt, roundcorner=4pt,linecolor=boxcolor,innerleftmargin=2pt,
innerrightmargin=6pt,innertopmargin=6pt,innerbottommargin=6pt]{annotationbox}
\renewcommand{\cite}{\shortcite}

\title{Grounding Language for Transfer in Deep Reinforcement Learning}

\author{\name Karthik Narasimhan \email karthikn@cs.princeton.edu \\
		\addr Department of Computer Science \\
       Princeton University \\
       35 Olden Street, Princeton, NJ 08540 USA \\ \\
       \name Regina Barzilay \email regina@csail.mit.edu \\
       \addr Computer Science and Artificial Intelligence Laboratory \\
       Massachusetts Institute of Technology \\
       32 Vassar Street, Cambridge, MA  02139 USA \\ \\
       \name Tommi Jaakkola \email tommi@csail.mit.edu \\
       \addr Computer Science and Artificial Intelligence Laboratory \\
       Massachusetts Institute of Technology \\
       32 Vassar Street, Cambridge, MA  02139 USA
       }

\date{}

\begin{document}
\maketitle

\input{abstract}

\input{introduction}
\input{relatedwork}

\input{formulation}
\input{model}

\input{experiments}
\input{results}
\input{conclusions}

\section*{Acknowledgements}
This work was done while Karthik Narasimhan was affiliated with MIT. We thank Adam Fisch, Victor Quach, and members of the MIT NLP group for their comments on earlier drafts of this paper.

\bibliography{references}

\bibliographystyle{theapa}

\end{document}


\maketitle

\section{Experimental Settings}
For all models, we used $\gamma = 0.8$, $\mathcal{D} = 250\text{k}$. We used a learning rate of $10^{-4}$, annealed linearly to $5 \times 10^{-5}$ and the \emph{Adam}~\cite{kingma2014adam} optimization scheme. The minibatch size is set to 32 and $\epsilon$ is annealed from 1 to 0.1 in the source tasks and set to 0.1 in the target tasks. For the value iteration module (VIN), we experimented with different levels of recurrence, $k \in \{1,3,5,10\}$ and found $k=3$ or $k=5$ to work best.\footnote{We still observed transfer gains with all $k$ values.} For the DQN, we used two convolutional layers followed by a single fully connected layer, with ReLU non-linearities. For the composition function $g$ in our model, we simply used component-wise addition. 

\bibliography{references}

\bibliographystyle{emnlp_natbib}

%% file: abstract.tex
\begin{abstract}
In this paper, we explore the utilization of natural language to drive transfer for reinforcement learning (RL). Despite the wide-spread application of deep RL techniques, learning generalized policy representations that work across domains remains a challenging problem. We demonstrate that textual descriptions of environments provide a compact intermediate channel to facilitate effective policy transfer. Specifically, by learning to ground the meaning of text to the dynamics of the environment such as transitions and rewards, an autonomous agent can effectively bootstrap policy learning on a new domain given its description. We employ a model-based RL approach consisting of a differentiable planning module, a model-free component and a factorized state representation to effectively use entity descriptions. Our model outperforms prior work on both transfer and multi-task scenarios in a variety of different environments. For instance, we achieve up to 14\% and 11.5\% absolute improvement over previously existing models in terms of average and initial rewards, respectively.
\end{abstract}

%% file: introduction.tex
\section{Introduction}
\label{sec:introduction}


Deep reinforcement learning has emerged as a method of choice for many control applications, ranging from computer games~\cite{mnih2015dqn,silver2016mastering} to robotics~\fullcite{levine2016end}. However, the success of this approach depends on a substantial number of interactions with the environment during training, easily reaching millions of steps~\cite{nair2015massively,mnih2016asynchronous}. Moreover, given a new task, even a related one, this training process has to be performed from scratch. This inefficiency has motivated recent work in learning universal policies that can generalize across related tasks~\fullcite{schaul2015universal}, as well as other transfer approaches~\fullcite{parisotto2016actor,rajendran20172t}.  In this paper, we explore transfer methods that use \emph{text descriptions} to facilitate policy generalization across tasks.


\begin{figure}[!t]
\minipage{\linewidth}
$\vcenter{\hbox{\includegraphics[width=0.55\linewidth]{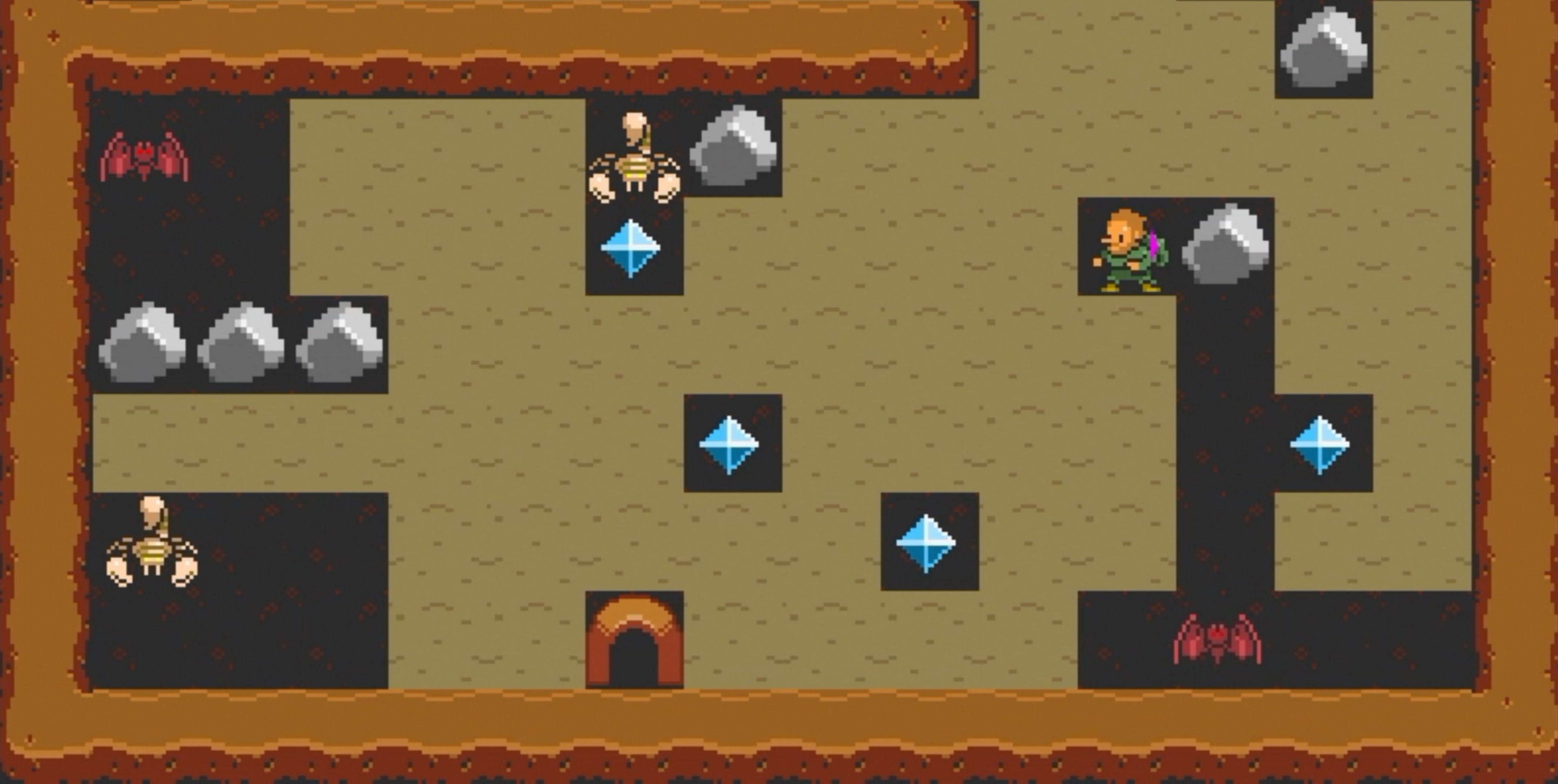}}}$ \hfill
$\vcenter{\hbox{\includegraphics[width=0.4\linewidth]{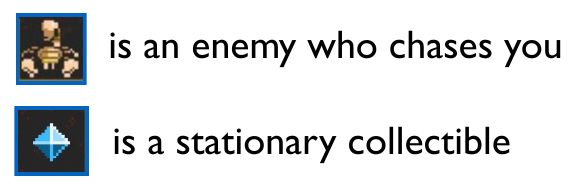}}}$
\endminipage\\

\minipage{\linewidth}
\vspace{0.3cm}
$\vcenter{\hbox{\includegraphics[width=0.55\linewidth]{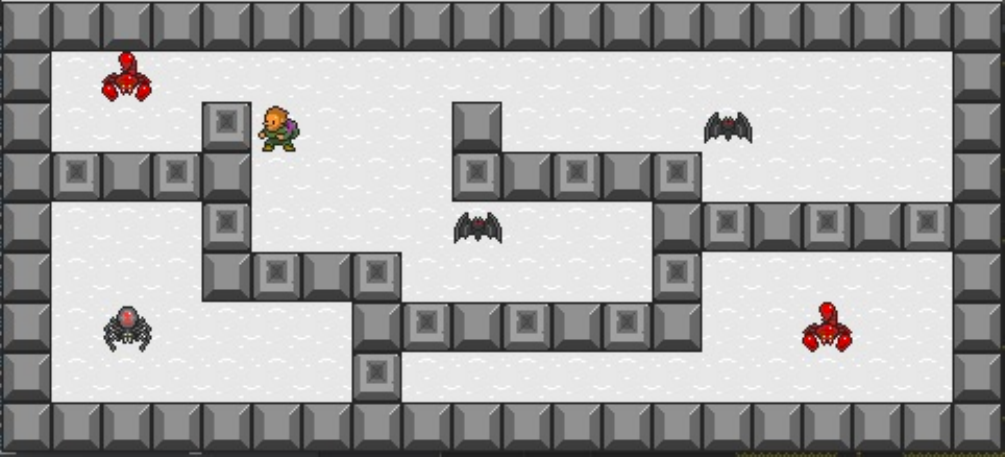}}}$\hfill
$\vcenter{\hbox{\includegraphics[width=0.4\linewidth]{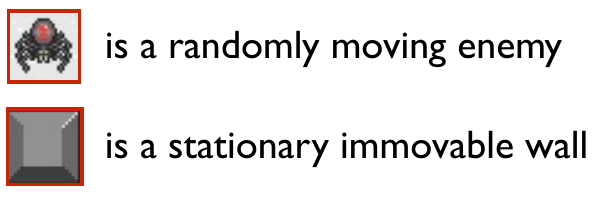}}}$
\endminipage 

\caption{Examples of two different game environments, Boulderchase \textbf{(top)} and Bomberman \textbf{(bottom)}. Each domain has text descriptions (collected using Amazon Mechanical Turk) associated with specific entities, describing characteristics such as movement and interactions with the player's avatar. Note how certain pairs of entities across games share certain properties. For instance, the scorpion in Boulderchase and the spider in Bomberman are both mobile entities.}

	\label{fig:example}
\end{figure}

As an example, consider the game environments in Figure~\ref{fig:example}. The two games -- \emph{Boulderchase} and \emph{Bomberman} 
--  differ in their layouts and entity types. However, the high-level behavior of most entities in both games is similar. For instance, the \emph{scorpion} in Boulderchase (top) is a moving entity which the agent has to avoid, similar to the \emph{spider} in Bomberman (bottom). Though this similarity is clearly reflected in the text descriptions in Figure~\ref{fig:example}, it may take multiple environment interactions to discover.  Therefore, exploiting these textual clues could help an autonomous agent understand this connection more effectively, leading to faster policy learning.




To test this hypothesis, we consider multiple environments augmented with textual descriptions. 
These descriptions provide a short overview of objects and their modes of interaction in the environment.\footnote{We do not require that every object to have an associated description.} They do not describe control strategies, which were commonly used in prior work on grounding~\fullcite{vogel2010learning,branavan2012learning}. Instead, they specify the dynamics of the environments, which are more conducive to cross-domain transfer.




In order to effectively use this type of information, we employ a model-based reinforcement learning approach. Typically, representations of the environment learned by these approaches are inherently domain-specific. We address this issue by using natural language as an implicit intermediate channel for transfer. Specifically, our model learns to map text descriptions to transitions and rewards in an environment, a capability that speeds up learning in unseen domains. We induce a two-part representation for the input state that generalizes over domains, incorporating both domain-specific information and textual knowledge. This representation is utilized by an action-value function, parametrized as a single deep neural network with a differentiable value iteration module~\fullcite{tamar2016value}. The entire model is trained end-to-end using rewards from the environment. 


We evaluate our model on several game worlds from the GVGAI framework~\fullcite{perez2016general}. In our main evaluation scenario of transfer learning, an agent is trained on a set of source tasks and its learning performance is evaluated on a different set of target tasks. Across multiple evaluation metrics, our method consistently outperforms several baselines and an existing transfer approach for deep reinforcement learning called Actor Mimic~\cite{parisotto2016actor}. For instance, our model achieves up to 14\% higher average reward and up to 11.5\% higher initial reward - two key metrics used to evaluate transfer learning~\fullcite{taylor2009transfer}.  We also demonstrate our model's improved performance on a multi-task setting where learning is simultaneously performed on multiple environments.

The rest of this paper is organized as follows. Section~\ref{sec:related_work} summarizes related work on grounding and transfer for reinforcement learning; Section~\ref{sec:framework} provides an overview of the framework we use; Section~\ref{sec:model} describes our model architecture and its various components; Section~\ref{sec:experiments} details the experimental setup, and Section~\ref{sec:results} contains our empirical results and analysis. We conclude and discuss some future directions for research in Section~\ref{sec:conclusions}. Code for the experiments in this paper is available at \url{https://github.com/karthikncode/Grounded-RL-Transfer}.

%% file: relatedwork.tex
\section{Related Work}
\label{sec:related_work}
We now provide a brief overview of related work in the areas of language grounding and transfer for reinforcement learning.
%
%

\subsection{Grounding Language in Interactive Environments}
In recent years, there has been increasing interest in systems that can utilize textual knowledge to learn control policies. Such applications include interpreting help documentation~\fullcite{branavan2010reading}, instruction following~\fullcite{vogel2010learning,kollar2010toward,artzi2013weakly,matuszek2013learning,Andreas15Instructions} and learning to play computer games~\fullcite{branavan2011nonlinear,branavan2012learning,narasimhan2015language,he2016deep}. In all these applications, the models are trained and tested on the same domain.

Our work represents two departures from prior work on grounding. First, rather than optimizing control performance for a single domain,
we are interested in the multi-domain transfer scenario, where language 
descriptions drive generalization. Second, prior work used text in the form of strategy advice to directly learn the policy. Since the policies are typically optimized for a specific task, they may be harder to transfer across domains. Instead, we utilize text to bootstrap the induction of the environment dynamics, moving beyond task-specific strategies. 


Another related line of work consists of systems that learn spatial and topographical maps of the environment for robot navigation using natural language descriptions~\fullcite{walter2013learning,hemachandra2014learning}. These approaches use text mainly containing appearance and positional information, and integrate it with other semantic sources (such as appearance models) to obtain more accurate maps. In contrast, our work uses language describing the dynamics of the environment, such as entity movements and interactions, which 
is complementary to static positional information received through state observations. Further, our goal is to help an agent learn policies that generalize over different stochastic domains, while their works consider a single domain.


\subsection{Transfer in Reinforcement Learning}
Transferring policies across domains is a challenging problem in reinforcement learning~\fullcite{konidaris2006framework,taylor2009transfer}. The main hurdle lies in learning a good mapping between the state and action spaces of different domains to enable effective transfer. Most previous approaches have either explored skill transfer~\fullcite{konidaris2007building,konidaris2012transfer} or value function/policy transfer~\fullcite{liu2006value,taylor2007transfer,taylor2007cross}. There have also been attempts at model-based transfer for RL~\fullcite{taylor2008transferring,nguyen2012transferring,gavsic2013pomdp,wang2015learning,joshi2018cross} but these methods either rely on hand-coded inter-task mappings for state and actions spaces or require significant interactions in the target task to learn an effective mapping. Our approach doesn't use any explicit mappings and can learn to predict the dynamics of a target task using its descriptions.


A closely related line of work concerns transfer methods for deep reinforcement learning. \citeA{parisotto2016actor}  train a deep network to mimic pre-trained experts on source tasks using policy distillation. The learned parameters are then used to initialize a network on a target task to perform transfer. Rusu et al.~\citeyear{rusu2016progressive} facilitate transfer by freezing parameters learned on source tasks and adding a new set of parameters for every new target task, while using both sets to learn the new policy. Work by Rajendran et al.~\citeyear{rajendran20172t} uses attention networks to selectively transfer from a set of expert policies to a new task. \textcolor{black}{Barreto et al.~\citeyear{barreto2017successor} use features based on successor representations~\fullcite{dayan1993improving} for transfer across tasks in the same domain. Kansky~et~al.~\citeyear{kansky2017schema} learn a generative model of causal physics in order to help zero-shot transfer learning.} Our approach is orthogonal to all these directions since we use text to bootstrap transfer, and can potentially be combined with these methods to achieve more effective transfer. 

\textcolor{black}{There has also been prior work on zero-shot policy generalization for tasks with input goal specifications. \fullciteA{schaul2015universal} learn a universal value function approximator that can generalize across both states and goals. \fullcite{andreas2016modular} use policy sketches, which are annotated sequences of subgoals, in order to learn a hierarchical policy that can generalize to new goals. \fullciteA{oh2017zero} investigate zero-shot transfer for instruction following tasks, aiming to generalize to unseen instructions in the same domain. The main departure of our work compared to these is in the use of environment descriptions for generalization across domains rather than generalizing across text instructions.}

Perhaps closest in spirit to our hypothesis is the recent work by~\fullcite{harrison2017guiding}. Their approach makes use of paired instances of text descriptions and state-action information from human gameplay to learn a machine translation model. This model is incorporated into a policy shaping algorithm to better guide agent exploration. Although the motivation of language-guided control policies is similar to ours, their work considers transfer across tasks in a single domain, and requires human demonstrations to learn a policy.

\textcolor{black}{
\subsection{Using Task Features for Transfer}
The idea of using task features/dictionaries for zero-shot generalization has been explored previously in the context of image classification. \fullciteA{kodirov2015unsupervised} learn a joint feature embedding space between domains and also induce the corresponding projections onto this space from different class labels. 
\fullciteA{kolouri2018joint} learn a joint dictionary across visual features and class attributes using sparse coding techniques. \fullciteA{romera2015embarrassingly} model the relationship between input features, task attributes and classes as a linear model to achieve efficient yet simple zero-shot transfer for classification. \fullciteA{socher2013zero} learn a joint semantic representation space for images and associated words to perform zero-shot transfer.}

\textcolor{black}{
Task descriptors have also been explored in zero-shot generalization for control policies. \fullciteA{sinapov2015learning} use task meta-data as features to learn a mapping between pairs of tasks. This mapping is then used to select the most relevant source task to transfer a policy from. \fullciteA{isele2016using} build on the ELLA framework~\fullcite{ruvolo2013ella,ammar2014online}, and their key idea is to maintain two shared linear bases across tasks -- one for the policy ($L$) and the other for task descriptors ($D$). Once these bases are learned on a set of source tasks, they can be used to predict policy parameters for a new task given its corresponding descriptor. 
In these lines of work, the task features were either manually engineered or directly taken from the underlying system parameters defining the dynamics of the environment. In contrast, our framework only requires access to crowd-sourced textual descriptions, alleviating the need for expert domain knowledge.}




%% file: formulation.tex
\section{General Framework}
\label{sec:framework}
Our goal in this work is to demonstrate the utility of natural language descriptions in assisting policy transfer across domains. In this section, we first describe our environment setup and the general framework of our approach. The details of our model and algorithm follow in Section~\ref{sec:model}.

\subsection{Environment Setup} 
We model a single environment as a Markov Decision Process (MDP),  $E = \langle S, A, T, R, O, Z \rangle$. Here, $S$ is the state space, and $A$ is the set of actions available to the agent. In this work, we consider every state $s \in S$ to be a 2-dimensional grid of size $m \times n$, with each cell containing an entity symbol $o \in O$.\footnote{In our experiments, we relax this assumption to allow for multiple entities per cell, but for ease of description, we shall assume a single entity per cell. The assumption of 2-D worlds can also be easily relaxed to generalize our model to other situations.} $T$ is the transition distribution over all possible next states $s'$ conditioned on the agent choosing action $a$ in state $s$. $R$ determines the reward provided to the agent at each time step. The agent does not have access to the true $T$ and $R$ of the environment. Each domain also has a goal state $s_g \in S$ which determines when an episode terminates. Finally, $Z$ is the complete set of text descriptions provided to the agent for this particular environment. 

\subsection{Reinforcement Learning (RL)}
The goal of an autonomous agent is to maximize cumulative reward obtained from the environment. A traditional way to achieve this is by learning an action value function $Q(s,a)$ through reinforcement. The \emph{Q-function} predicts the expected future reward for choosing action~$a$ in state~$s$. A straightforward policy then is to simply choose the action that maximizes the $Q$-value in the current state: 

\begin{dmath*}
\pi(s) = \argmax_a Q(s,a)
\end{dmath*}

If we also make use of the descriptions, we have a text-conditioned policy: 
\begin{dmath}
\pi(s, Z) = \argmax_a Q(s, a, Z)
\end{dmath} 

A successful control policy for an environment will contain both knowledge of  the environment dynamics and the capability to identify goal states. While the latter is task-specific, the former characteristic is more useful for learning a general policy that transfers to different domains. Based on this hypothesis, we employ a model-aware RL approach that can learn the dynamics of the world while estimating the optimal $Q$. Specifically, we make use of \emph{Value Iteration (VI)}~\cite{sutton1998introduction}, an algorithm based on dynamic programming. The update equations for value iteration in our setup are:
\begin{align}
Q^{(n+1)}(s, a, Z) &= \sum_{s' \in S} T(s' | s, a, Z) [ R(s', Z) + \gamma V^{(n)}(s', Z) ]  \nonumber \\
V^{(n+1)}(s, Z) &= \max_a Q^{(n+1)}(s,a, Z) 
\label{eq:vi}
\end{align}
where $\gamma$ is a discount factor and $n$ is the iteration number. The updates require an estimate of $T$ and $R$, which the agent must obtain through exploration of the environment.


\subsection{Text Descriptions}
Estimating the dynamics of the environment from interactive experience can require a significant number of samples. Our main hypothesis is that if an agent can derive information about the dynamics from text descriptions, it can determine $T$ and $R$ faster and more accurately. 

For instance, consider the sentence \emph{``Red bat that moves horizontally, left to right''}. This talks about the movement of a third-party entity (\emph{bat}), independent of the agent's goal. Provided the agent can learn to interpret this sentence, it can then infer the direction of movement of a different entity (e.g. \emph{``A tan car moving slowly to the left''}) in a different domain. Further, this inference is useful even if the agent has a completely different goal. On the other hand, instruction-like text such as \emph{``Move towards the wooden door''} is highly context-specific and only relevant to domains that have the mentioned goal.

With this in mind, we provide the agent with text descriptions that collectively portray characteristics of the world. These descriptions are crowdsourced by asking humans to view gameplay videos and describe entities.  A single description talks about one particular entity in the world. The text contains (partial) information about the entity's movement and interaction with the player avatar. Each description is also aligned to its corresponding entity in the environment and not all entities may have a description.
Figure~\ref{fig:descriptions} provides some samples; more details on data collection and statistics are in Section~\ref{sec:experiments}. 

\begin{figure}
  \begin{annotationbox}
    \small
      \begin{itemize}
        \item Scorpion2: \emph{Red scorpion that moves up and down} 
        \item Alien3: \emph{This character slowly moves from right to left while having the ability to shoot upwards}
        \item Sword1: \emph{This item is picked up and used by the player for attacking enemies}
      \end{itemize}
  \end{annotationbox}
  \caption{Example text descriptions of entities in different environments, collected using Amazon Mechanical Turk. Turkers were shown videos of gameplay in the different environments and asked to describe each entity's behavior or role. Note that these sentences are not instructive, since they provide no direct information on how to act in the environment.}
  \label{fig:descriptions}
\end{figure}

\subsection{Transfer for RL}
In order to test our grounding hypothesis, we consider learning across multiple environments. Specifically, an agent can learn to ground language semantics in an environment $E_1$ and then we can test its understanding capability by placing it in a new unseen domain, $E_2$. The agent can obtain unlimited experience in $E_1$, and after convergence of its policy, it is allowed to interact with and learn a policy for $E_2$. We do not provide the agent with any explicit mapping between different entities or goals across domains, either directly or through the text. For instance, even though the boulders in \emph{Boulderchase} are impassable objects just like the walls in \emph{Bomberman}~\ref{fig:example}, the agent does not have access to a mapping between these entities. In this setup, the agent's goal is to re-utilize information obtained through its interactions in $E_1$ to learn more efficiently in $E_2$.

%% file: model.tex
\section{Model}
\label{sec:model}

Grounding language for policy transfer across domains requires a model that meets two needs. First, it must allow for a flexible representation that fuses information from both state observations and text descriptions. This representation should capture the compositional nature of language while mapping linguistic semantics to characteristics of the world. Second, the model must have the capability to learn an accurate prototype  of the environment (i.e. transitions and rewards) using only interactive feedback. Overall, the model must enable an agent to map text descriptions to environment dynamics; this allows it to predict transitions and rewards in a completely new world, without requiring substantial interaction.

To this end, we propose a neural architecture consisting of two main components: (1) a \emph{representation generator} ($\phi$), and (2) a \emph{value iteration network} (VIN)~\cite{tamar2016value}. First, the representation generator takes a state observation and a set of text descriptions as input and produces a tensor output, capturing  information essential for decision-making. Then, the VIN module implicitly encodes the value iteration computation (Eq.~\ref{eq:vi}) into a recurrent network with convolutional modules, producing an action-value function using the previously constructed tensor representation as input. Together, both modules form an end-to-end differentiable network that can be trained using gradient back-propagation.  
 
\subsection{Representation Generator}

\begin{figure}[t]
\centering
 \includegraphics[width=0.8\linewidth]{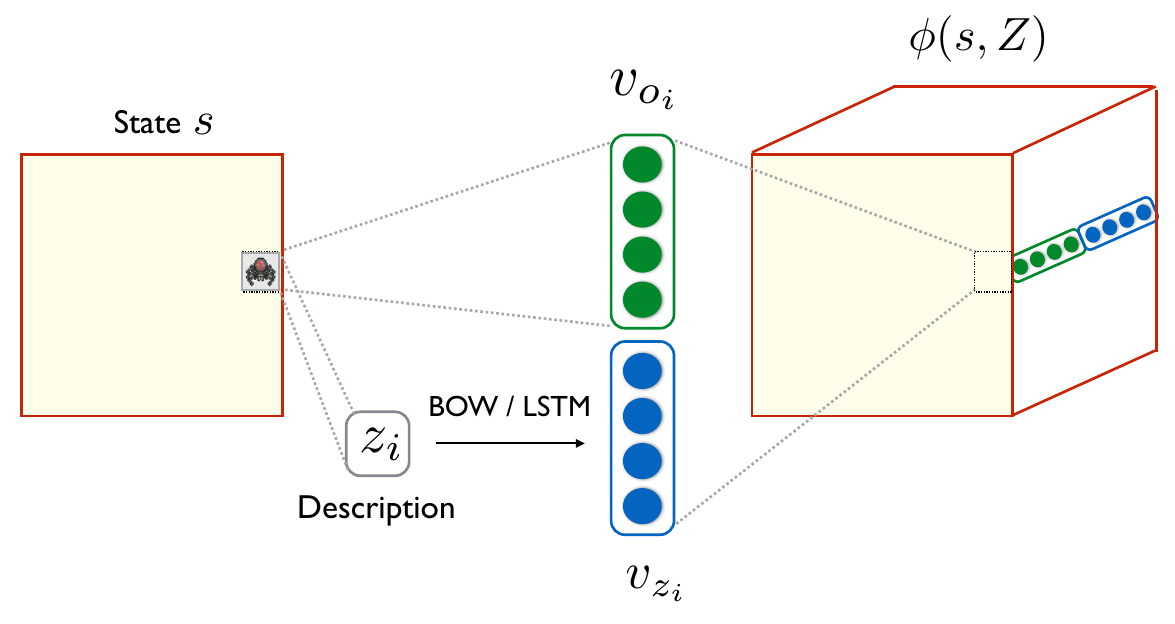}
\caption{Representation generator combining both object-specific and description-informed vectors for each entity. Each cell in the input state (2-D matrix) is converted to a corresponding real-valued vector, resulting in a 3-D tensor output. The two-part representation allows us to exploit partial/noisy information from text while also learning other aspects of the environment dynamics directly through interaction.} 
	\label{fig:transfer-state}
\end{figure}

The main purpose of this module (Figure~\ref{fig:transfer-state}) is to fuse together information from two inputs -- the state, and the text specifications. An important consideration, however, is the ability to handle partial or incomplete text descriptions, which may not contain all the particulars of an entity. Thus, we would like to incorporate useful information from the text, yet, not rely on it completely. This motivates us to use a representation that is factorized over the two input modalities.

\color{black}
Formally, given a state matrix $s$ of size $m \times n$ and a set of text descriptions $Z$, the module produces a tensor $\phi(s, Z)$. Recall that each cell in state $s$ is occupied by a single entity. Consider one such cell containing an entity $o_i$, with a corresponding description $z_i$ (if available). The representation generator performs two operations:
\begin{enumerate}
\item First, it generates an entity-specific vector $v_{o_i}$ of dimension $d$. This vector is initialized arbitrarily and learned using rewards received by the agent in the environment. One can view this operation as an `object embedding', similar to the notion of a word embedding~\fullcite{mikolov2013efficient}.
\item Second, the description $z_i$ is converted into a continuous valued vector  \textbf{$v_{z_i}$} (also of dimension $d$). This can be achieved in several different ways, but in this work, we experiment with using an LSTM recurrent neural network~\cite{hochreiter1997long} and a mean bag-of-words (BOW) approach, which entails taking the average over word vectors corresponding to each word in the description. 
\end{enumerate}
Both these sets of parameters (including the embeddings and LSTM weights) are all initialized at random and learned through reinforcement on the source tasks.

The two vectors, $v_{o_i}$ and  $v_{o_z}$,  are then concatenated to produce a single representation for this cell: $\phi_i = [v_{o_i};v_{z_i}]$. Performing the same operations over all cells of the state results in a tensor $\phi(s,Z)$ with dimensions $m \times n \times 2d$ for the entire state.\footnote{$d$ is a hyperparameter choice here. Also, one can have vectors $v_{o_i}$ and $v_{z_i}$ of different dimensions, say $d_1$ and $d_2$, if necessary. We use the same dimensionality for simplicity.} For cells with no entity (i.e. empty space), $\phi_i$ is simply a zero vector, and for entities without a description, $v_{z_i} = \vec{0}$.  Figure~\ref{fig:transfer-state} illustrates this module.

This decomposition into $v_o$ and $v_z$ allows us to learn policies based on both the ID of an object and its described behavior in text. This enables the model to retain knowledge of observed entities while being adaptable to new entities. For instance, if a new environment contains some previously seen entities,\footnote{Note that this is just a possible situation our model can handle. The different domains we consider in this work have no entity symbol overlap.} the agent can reuse the learned representations directly based on their symbols. For completely new entities (with unseen IDs), the model can form useful representations using their text descriptions. 

\color{black}
\subsection{Value Iteration Network}

\begin{figure}[!t]
\centering
  \includegraphics[width=\linewidth]{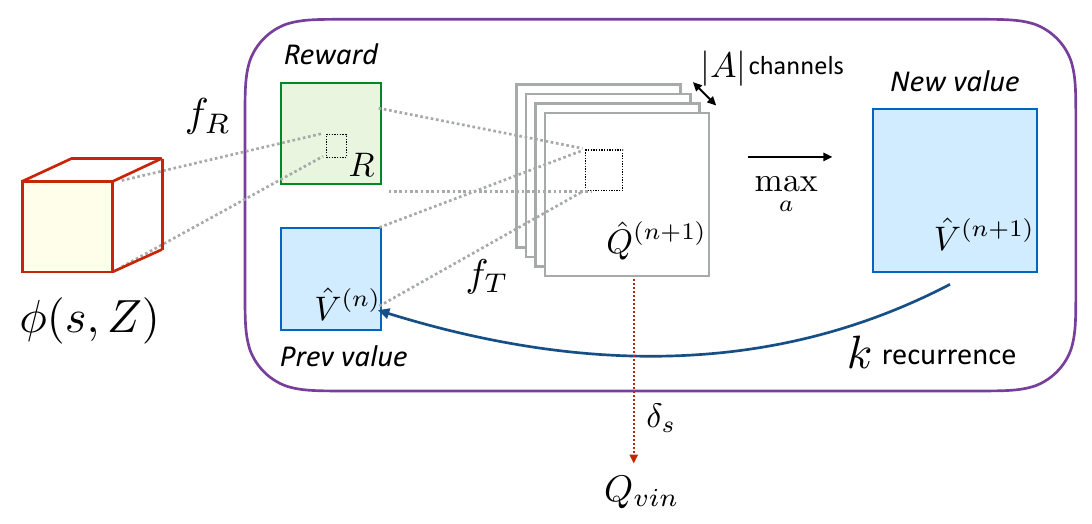}
\caption{Value iteration network (VIN) module to compute $Q_{vin}$ from $\phi(s, Z)$. The module approximates the value iteration computation using neural networks to predict reward and value maps, arranged in a recurrent fashion. Functions $f_T$ and $f_R$ are implemented using convolutional neural networks (CNNs). $\delta_s$ is a selection function to pick out a single Q-value (at the agent's current location) from the output Q-value map $\hat{Q}^{(k)}$.}
	\label{fig:transfer-model}
\end{figure}

For a model-based RL approach to this task, we require some means to estimate $T$ and $R$ of an environment. One way to achieve this is by explicitly using predictive models for both functions and learning these through transitions experienced by the agent. These models can then be used to estimate the optimal $Q$ using equation~\ref{eq:vi}. However, this pipelined approach would result in  errors propagating through the different stages of predictions.
 
A value iteration network (VIN)~\cite{tamar2016value} abstracts away explicit computation of $T$ and $R$ by directly predicting the outcome of value iteration (Figure~\ref{fig:transfer-model}), thereby avoiding the aforementioned error propagation. In this model, the VI computation is mimicked by a recurrent network with two key operations at each step. First, to compute $Q$, we have two functions -- $f_R$ and $f_T$. $f_R$ is a reward predictor that operates on $\phi(s, Z)$ while $f_T$ uses the output of $f_R$ and any previous $V$ to predict $Q$. Both functions are parametrized as convolutional neural networks (CNNs),\footnote{Other parameterizations are possible for different input types, as noted by~\citeA{tamar2016value}.} to suit our tensor representation $\phi$. Subsequently, in the second operation, the network employs max pooling over the action channels in the $Q$-value map produced by $f_T$ to obtain $V$. 
The value iteration computation (from Eq.~\ref{eq:vi}) can thus be approximated as:
\begin{dmath}
\hat{Q}^{(n+1)}(s, a, Z) = f_T \Big(f_R(\phi(s,Z), a; \theta_R), \hat{V}^{(n)}(s, Z); \theta_T \Big) 
\end{dmath} \vspace{-0.1cm}
\begin{dmath}
\hat{V}^{(n+1)}(s, Z) = \max_a \hat{Q}^{(n+1)}(s,a, Z)
\end{dmath}
Note that while the VIN operates on $\phi(s,Z)$, we write $\hat{Q}$ and $\hat{V}$ in terms of the original state input $s$ and text $Z$, since these are independent of our chosen representation.

The outputs of both CNNs are real-valued tensors. The output of $f_R$ has the same dimensions as the input state  ($m \times n$), while $f_T$ produces $\hat{Q}^{(n+1)}$ as a tensor of dimension $m \times n \times |A|$, where $|A|$ is the number of actions available to the agent. A key point here is that the model produces $Q$ and $V$ values for each cell of the input state matrix, assuming the agent's position to be that particular cell. The convolution filters help capture information from neighboring cells in our state matrix, which act as approximations for $V^{(n)}(s', Z)$. The parameters of the CNNs, $\theta_R$ and $\theta_T$, approximate $R$ and $T$, respectively. See the work of  Tamar~et~al.~\citeyear{tamar2016value} for a more detailed discussion.

The recursive computation of traditional value iteration~(Eq.~\ref{eq:vi}) is captured by employing the CNNs in a recurrent fashion for $k$ steps.\footnote{$k$ is a model hyperparameter.} Intuitively, larger values of $k$ imply a larger field of neighbors influencing the Q-value prediction for a particular cell as the information propagates longer. The final output of this recurrent computation, $\hat{Q}^{(k)}$, is a 3-D tensor of size $m \times n \times |A|$. However, since we need a policy only for the agent's current location, we use an appropriate selection function $\delta_s$, which reduces this Q-value map to a single set of action values for the agent's location:
\begin{align}
Q_{vin}(s, a, Z; \Theta_1) &=  \delta_s (\hat{Q}^{(k)} (s, a, Z))
\end{align}
This is simply an indexing operation performed on $\hat{Q}^{(k)}$ to retrieve the $|A|$-dimensional vector from the cell corresponding to the agent's location in state $s$.

\subsection{Final Prediction} 
Games exhibit complex dynamics, which are challenging to capture precisely, especially for long-term prediction. VINs approximate the dynamics implicitly via learned convolutional operations. It is thus likely that the estimated $Q_{vin}$ values are most helpful for short-term planning that corresponds to a limited number of iterations $k$. Therefore, we need to complement these `local' Q-values with estimates based on a more global view.

To this end, following the VIN specification by~\citeA{tamar2016value}, our architecture also contains a model-free action-value function, implemented as a deep Q-network (DQN)~\cite{mnih2015dqn}. This network provides a Q-value prediction -- $Q_{r} (s,a,Z; \Theta_2)$ -- which is combined with $Q_{vin}$ using a composition function $g$:\footnote{Although $g$ can also be learned, we use component-wise addition in our experiments.}
\begin{dmath}
Q(s, a, Z; \Theta) = g(Q_{vin}(s, a, Z; \Theta_1), Q_{r}(s, a, Z; \Theta_2))
\label{eq:final-q}
\end{dmath}
The fusion of our model components enables our agent to establish the connection between input text descriptions, represented as vectors, and the environment's transitions and rewards, encoded as VIN parameters. In a new domain, the model can produce a reasonable policy using corresponding text, even before receiving any interactive feedback.


\subsection{Parameter Learning}
Our entire model is end-to-end differentiable. We perform updates derived from the Bellman equation~\cite{sutton1998introduction}:
\begin{dmath}
	{Q_{i+1}(s,a, Z) = \mathrm{E}[r + \gamma \max_{a'} Q_i(s',a', Z) \mid s, a]} 
\label{eq:transfer-bellman-update}
\end{dmath}
where the expectation is over all transitions from state $s$ with action $a$ and $i$ is the update number. To learn our parametrized Q-function (the result of Eq.~\ref{eq:final-q}), we can use backpropagation through the network to minimize the following loss:
\begin{dmath}
	{\mathcal{L}(\Theta_i) = \mathrm{E}_{\hat{s},\hat{a}}  [ (y_i - Q(\hat{s}, \hat{a}, Z ; \Theta_i))^2 ]}
\label{eq:transfer-loss}
\end{dmath}
where $ {y_i = r + \gamma \max_{a'} Q (s',a', Z; \Theta_{i-1})}$ is the target Q-value with parameters $\Theta_{i-1}$ fixed from the previous iteration.  We employ an experience replay memory $\mathcal{D}$ to store transitions~\cite{mnih2015dqn}, and periodically perform updates with random samples from this memory. We use an $\epsilon$-greedy policy~\cite{sutton1998introduction} for exploration.  

 \begin{algorithm}[t]
\caption{\textsc{Multitask\_Train ($\mathcal{E}$)}}
\label{alg:transfer-train}
\begin{algorithmic}[1]
\State Initialize parameters $\Theta$ and experience replay $\mathcal{D}$
\For {$ k = 1,M $}  \Comment{New episode}
	\State Choose next environment $E_k \in \mathcal{E}$
	\State Initialize $E_k$; get start state $s_1 \in S_k$	
	\For {$ t = 1, N $}   \Comment{New step}
    	\State Select $a_t \sim \textsc{eps-greedy}(s_t, Q_\Theta, Z_k, \epsilon)$		
		\State Execute action $a_t$, observe reward $r_t$ and new state $s_{t+1}$
		\State $\mathcal{D} = \mathcal{D} \cup (s_t, a_t, r_t, s_{t+1}, Z_k)$
		\State Sample mini batch $(s_j, a_j, r_j, s_{j+1}, Z_k) \sim \mathcal{D}$
		\State Perform gradient descent on loss $\mathcal{L}$ to update $\Theta$
		\If { $s_{t+1}$ is terminal}
			\textbf{break}
		\EndIf
	\EndFor
\EndFor
\State Return $\Theta$
\end{algorithmic}
\end{algorithm}

\begin{algorithm}[t]
\caption{\textsc{eps-greedy} ($s,Q,Z,\epsilon$)}
\label{alg:transfer-eps-greedy}
\begin{algorithmic}[1]
\If {$ random() < \epsilon $}			
			\State Return random action $a$
		\Else		        	
			\State Return  $\argmax_a~Q(s, a, Z)$  
		\EndIf
\State Return $\Theta$
\end{algorithmic}
\end{algorithm}

\subsection{Transfer Procedure}
The traditional transfer learning scenario often involves a single task in both source and target environments. To better test generalization and robustness of our methods, in this work we consider transfer from \emph{multiple} source tasks to \emph{multiple} target tasks.
We first train a model to achieve optimal performance on the set of source tasks. All model parameters ($\Theta$) are shared between these tasks. The agent experiences one episode at a time, sampled from each environment in a round-robin fashion, along with the corresponding text descriptions. The parameters of the model are optimized using the reward-based feedback gathered across all these tasks. Algorithm~\ref{alg:transfer-train} details this multi-task training procedure.

After training converges, we use the learned parameters to initialize a model for tasks in the target domain. Specifically, all parameters of the VIN are replicated, while most weights of the representation generator are reused. Previously seen objects and words retain their learned entity-specific embeddings ($v_o$), whereas vectors for new objects/words in the target tasks are initialized randomly. Following this initialization, all parameters are then fine-tuned on the target tasks using the corresponding rewards, again with episodes sampled in a round-robin fashion.

%% file: experiments.tex
\section{Experimental Setup}
\label{sec:experiments}
We now detail our empirical setup including environments, text descriptions, evaluation metrics, baselines and model implementation details. Results follow in Section~\ref{sec:results}.

\subsection{Environments}
We perform experiments on a series of 2-D environments within the GVGAI framework~\cite{perez2016general}, which is used in an annual video game AI competition.\footnote{\url{http://www.gvgai.net/}} In addition to pre-specified games, the framework supports the creation of new games using the Py-VGDL description language~\cite{schaul2013video}. We use four different games to evaluate transfer and multitask learning: \emph{Freeway}, \emph{Bomberman}, \emph{Boulderchase} and \emph{Friends \& Enemies}. There are certain similarities between these games. For one, each game consists of a 16x16 grid with the player controlling a movable avatar with two degrees of freedom. Also, each domain contains other entities, both stationary and moving (e.g. diamonds, spiders), that can interact with the avatar. 

However, each game also has its own distinct characteristics. In \emph{Freeway}, the goal is to cross a multi-lane freeway while avoiding cars in the lanes. The cars move at various paces in either horizontal direction. \emph{Bomberman} and \emph{Boulderchase} involve the player seeking an exit door while avoiding enemies that either chase the player, run away or move at random. The agent also has to collect resources like diamonds and dig or place bombs to clear paths. These three games have five level variants each with different map layouts and starting entity placements. 

\emph{Friends \& Enemies} (F\&E) is a new environment we designed, with a larger variation of entity types. This game has a total of twenty different non-player entities, each with different types of movement and interaction with the player's avatar. For instance, some entities move at random while some chase the avatar or shoot bullets that the avatar must avoid. The objective of the player is to meet all friendly entities while avoiding enemies. For each game instance, four non-player entities are sampled from this pool and randomly placed in the grid. This makes F\&E instances significantly more varied than the previous three games.  We created two versions of this game: F\&E-1 and F\&E-2, with the sprites in F\&E-2 moving faster, making it a harder environment. Table~\ref{table:variations} contains all the different transfer scenarios we consider in our experiments. 

\begin{table}[h]
\centering
\begin{tabular}{  c  c  c  c } 
\textbf{Condition} & \textbf{Source} & \textbf{Target} & \textbf{\% vocab overlap}\\ \midrule
F\&E-1 $\rightarrow$ F\&E-2 & 7 & 3 & 100 \\
F\&E-1 $\rightarrow$ Freeway & 7 & 5 & 18.06 \\ 
Bomberman $\rightarrow$ Boulderchase & 5 & 5 & 19.6 \\ 
\end{tabular}
\caption{Statistics on source and target games for various transfer experiments. First two columns indicate the number of instances of each game, while the last column contains the percentage of overlap between the vocabularies of the corresponding text descriptions, collected using Amazon Mechanical Turk.}
\label{table:variations}
\end{table}

\subsection{Text Descriptions}
We collect textual descriptions using Amazon Mechanical Turk~\cite{mturk}. We provide annotators with sample gameplay videos of each game and ask them to describe specific entities in terms of their movement and interactions with the avatar. Since we ask the users to provide an independent account of each entity, we obtain \emph{descriptive} sentences as opposed to \emph{instructive} ones which inform the optimal course of action from the avatar's viewpoint.\footnote{Upon manual verification, we find less than 3\% of the obtained annotations to be instructive, i.e. containing text that explicitly instruct the agent on steps to take in order to achieve the goal.} 

We aggregated together four sets of descriptions, each from a different annotator, for every environment. This resulted in an average of around 36 unique sentences per domain, with \emph{F\&E} having the most: 78 sentences. Apart from lowercasing the text, we do not perform any extra pre-processing. Each description in an environment is aligned to one constituent entity. We also make sure that the entity names are not repeated across games (even for the same entity type). Table~\ref{table:stats} provides corpus-level statistics on the collected data and Figure~\ref{fig:descriptions} has sample descriptions.


\begin{table}[h]
\centering
\begin{tabular}{  c  c  } \toprule
Unique word types &  286 \\ 
Avg. words / sentence &  8.65\\
Avg. sentences / domain & 36.25 \\
Max sentence length & 22 \\ \bottomrule
\end{tabular}
\caption{Overall statistics of the text descriptions collected using Mechanical Turk.}
\label{table:stats}
\end{table}


\subsection{Evaluation Metrics}
We evaluate transfer performance using three metrics defined and employed in previous work~\cite{taylor2009transfer}:
\begin{itemize}[leftmargin=0.45cm]
\item \emph{Average Reward}, which is the area under the reward curve divided by the number of test episodes.
\item \emph{Initial performance}, which is the average reward over first 50k steps. 
\item \emph{Asymptotic performance}, which is the average reward over 50k steps after convergence. 
\end{itemize}

The first two metrics emphasize the speed at which a transfer method can enable learning on the target domain, while the last one evaluates its ability to achieve optimal performance on the task. An ideal method should provide gains on all three metrics.
For the multitask scenario, we consider the average and asymptotic reward only.  For each metric, we repeat experiments with nine different random seeds and report mean and standard deviation numbers.

\subsection{Baselines}
We explore several baseline models for empirical comparison. The different conditions we consider are:
\begin{itemize}[leftmargin=0.45cm]
\itemsep0em 
\item \textsc{no transfer}: A deep Q-network (DQN)~\cite{mnih2015dqn} is initialized randomly and trained from scratch on target tasks. This is the only case that does not use parameters transferred from source tasks.
\item \textsc{dqn}: A DQN is trained on source tasks and its parameters are transferred to target tasks. This model does not make use of text descriptions.
\item \textsc{text-dqn}: This is a DQN with our hybrid representation $\phi(s, Z)$, using the text descriptions. This is essentially a reactive-only version of our model, i.e. without the VIN planning module.
\item \textsc{amn}: The Actor-Mimic network is a recently proposed transfer method~\cite{parisotto2016actor} for deep RL. \textsc{amn} employs policy distillation to train a single network using expert policies previously learned on multiple different tasks. This network is then used to initialize a model for a new task.\footnote{We only evaluate AMN on transfer since it does not perform online multitask learning and is not directly comparable.}
\item \textsc{vin}: A value iteration network is trained on the source tasks without making use of the text descriptions. This is effectively an ablation of our full model that only receives state observations as input.
\end{itemize}

\subsection{Implementation Details}
We now provide details on our model implementations. For all models, we set $\gamma = 0.8$, $|\mathcal{D}| = 250\text{k}$, and the embedding size $d = 10$. We used the \emph{Adam}~\cite{kingma2014adam} optimization scheme with a learning rate of $10^{-4}$, annealed linearly to $5 \times 10^{-5}$. The minibatch size was set to 32. $\epsilon$ was annealed from 1 to 0.1 in the source tasks and set to 0.1 in the target tasks. For the value iteration module (VIN), we experimented with different levels of recurrence, $k \in \{1,2,3,5\}$ and found $k=1$ or $k=3$ to work best.\footnote{We still observed transfer gains with all $k$ values.} For DQN, we used two convolutional layers followed by a single fully connected layer, with ReLU non-linearities. The CNNs in the VIN  had filters and strides of length 3. The CNNs in the model-free component used filters of sizes $\{4,2\}$ and corresponding strides of size $\{3,2\}$. All embeddings are initialized at random.\footnote{We also experimented with using pre-trained word embeddings for text but obtained equal or worse performance.}

%% file: results.tex
 \begin{table*}[!t]
\centering

\begin{tabular}{ c  c c c } \\
\multicolumn{4}{c}{\textbf{F\&E-1 $\rightarrow$ F\&E-2}} \\
\textit{Model} & \emph{Average} & \emph{Initial} & \emph{Asymptotic}  \\ \hline
\rule{0pt}{3ex} \textsc{no transfer}  & 0.88 (0.10) & -0.24 (0.09) & 1.46 (0.07) \\ \cdashline{1-4}[0.5pt/5pt] \rule{0pt}{3ex}
\textsc{dqn}    & 0.98 (0.14) & 0.36 (0.10) & 1.20 (0.09) \\
\emph{\textsc{text-dqn}}  & 0.93 (0.13) & 0.47 (0.33) & 1.21 (0.10) \\
\textsc{amn} (Actor Mimic)       & 1.22 (0.05) & 0.13 (0.10) & \textbf{1.64} (0.01)\\
\textsc{vin (3)}        & 1.14 (0.08) & 0.12 (0.16) & 1.49 (0.07)  \\
\cdashline{1-4}[0.5pt/5pt]
\rule{0pt}{3ex}
\textsc{text-vin (3)} & \textbf{1.32} (0.06) & \textbf{0.70} (0.22) & 1.50 (0.05) \\
\end{tabular}

\begin{tabular}{ c  c c c } \\
\multicolumn{4}{c}{\textbf{F\&E-1 $\rightarrow$ Freeway}} \\
\textit{Model} & \emph{Average} & \emph{Initial} & \emph{Asymptotic}  \\ \hline
\rule{0pt}{3ex} \textsc{no transfer}  & 0.22 (0.03) & -0.95 (0.10) & 0.82 (0.03) \\ \cdashline{1-4}[0.5pt/5pt] \rule{0pt}{3ex}
\textsc{dqn}    & 0.21 (0.16) & -0.78 (0.17) & 0.78 (0.11) \\
\emph{\textsc{text-dqn}}  & 0.33 (0.10) & -0.72 (0.17) & 0.83 (0.01) \\
\textsc{amn} (Actor Mimic)       & 0.08 (0.03) & -0.84 (0.04) & 0.75 (0.005)\\
\textsc{vin (1)}        & 0.59 (0.16) & -0.32 (0.57) & \textbf{0.85} (0.01)  \\
\cdashline{1-4}[0.5pt/5pt]
\rule{0pt}{3ex}
\textsc{text-vin (3)} & \textbf{0.73} (0.01) & \textbf{-0.01} (0.09) & \textbf{0.85} (0.01) 
\end{tabular}

\begin{tabular}{ c c c c } \\
\multicolumn{4}{c}{\textbf{Bomberman $\rightarrow$ Boulderchase}} \\
\textit{Model} & \emph{Average} & \emph{Initial} & \emph{Asymptotic}  \\ \hline
\rule{0pt}{3ex} 
\textsc{no transfer}       & 8.16 (0.79) & 2.88 (0.29) & 10.67 (1.37) \\ \cdashline{1-4}[0.5pt/5pt] \rule{0pt}{3ex}
\textsc{dqn}               & 7.30 (1.39) & 3.77 (0.45) & 9.24 (1.83) \\
\emph{\textsc{text-dqn}}   & 7.92 (0.64) & 3.44 (0.54) & 10.10 (1.69) \\
\textsc{amn} (Actor Mimic) & 5.58 (0.53) & 1.08 (0.38) & 8.66 (0.97) \\
\textsc{vin (3)}           & 9.84 (0.51) & 3.77 (0.44)  & \textbf{12.22} (0.48)  \\
\cdashline{1-4}[0.5pt/5pt]
\rule{0pt}{3ex}
\textsc{text-vin (3)}    & \textbf{11.17} (0.44) & \textbf{5.37} (0.78) & 12.08 (0.31)\\
\end{tabular}

\caption{Transfer learning results under the different metrics for different domains. Numbers in parentheses for \textsc{vin} and \textsc{text-vin} indicate the $k$ value for the best model. \textsc{text-} models make use of textual descriptions.  Numbers are averaged over 9 independent runs (3 source $\times$ 3 target); higher scores are better; bold indicates best numbers; standard deviation numbers are in parentheses. The max reward attainable (ignoring step penalties) in the target environments is $2.0$, $1.0$ and at least $25.0$ in F\&E, Freeway and Boulderchase, respectively. }
\label{table:results}
\end{table*}


\section{Results}
\label{sec:results}
We now present empirical evidence that demonstrates the effectiveness of our approach. We first begin by analyzing performance under the transfer condition, followed by the multi-task results and an analysis of the model.

\subsection{Transfer Performance}
Table~\ref{table:results} demonstrates that transferring policies positively assists learning in new domains. Our model, \textsc{text-vin} achieves superior performance to the baselines on average and initial rewards in all transfer conditions.

\subsubsection{Average Reward}
\begin{figure}[!h] 
\minipage{\textwidth}
\centering
\includegraphics[width=0.54\linewidth]{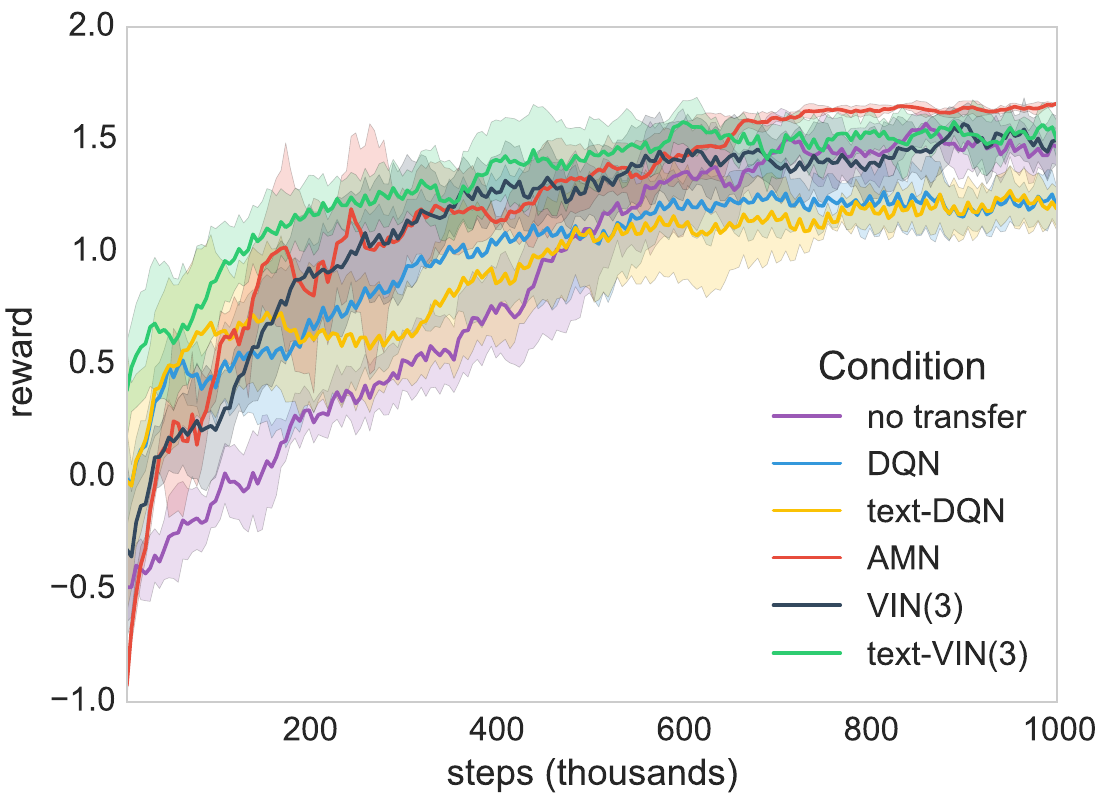}
\endminipage\hfill
\minipage{\textwidth}
\centering
\includegraphics[width=0.55\linewidth]{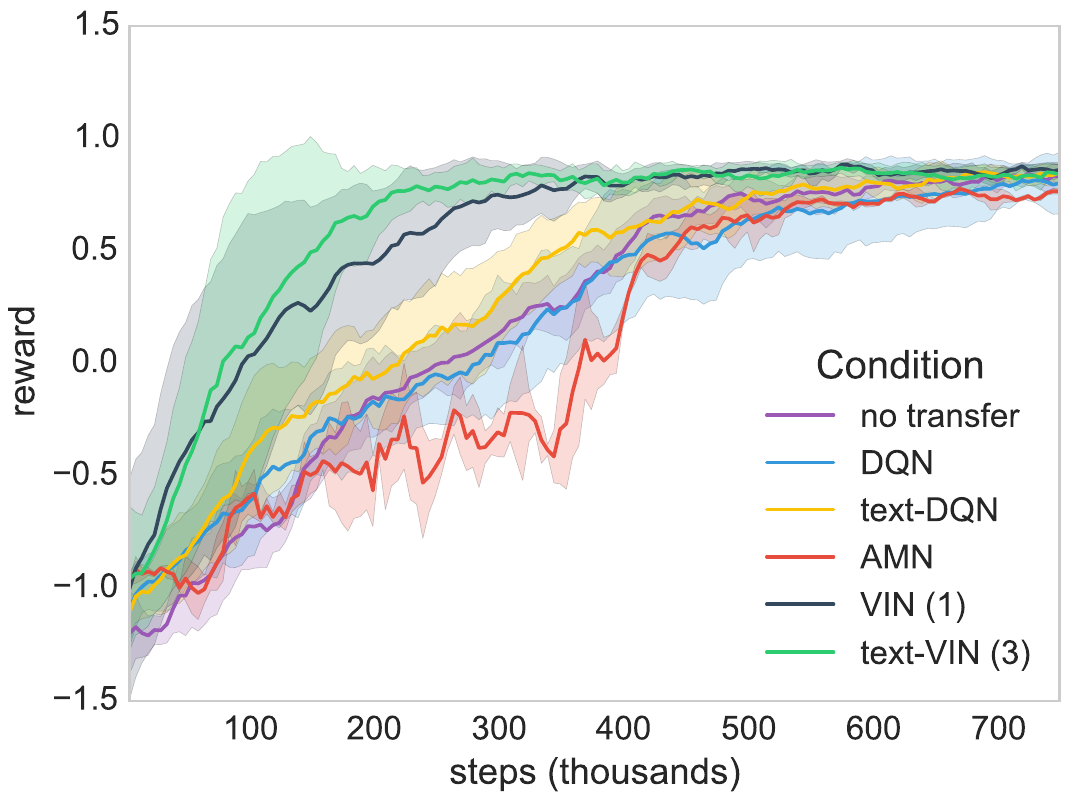}
\endminipage\hfill
\minipage{\textwidth}
\centering
\includegraphics[width=0.54\linewidth]{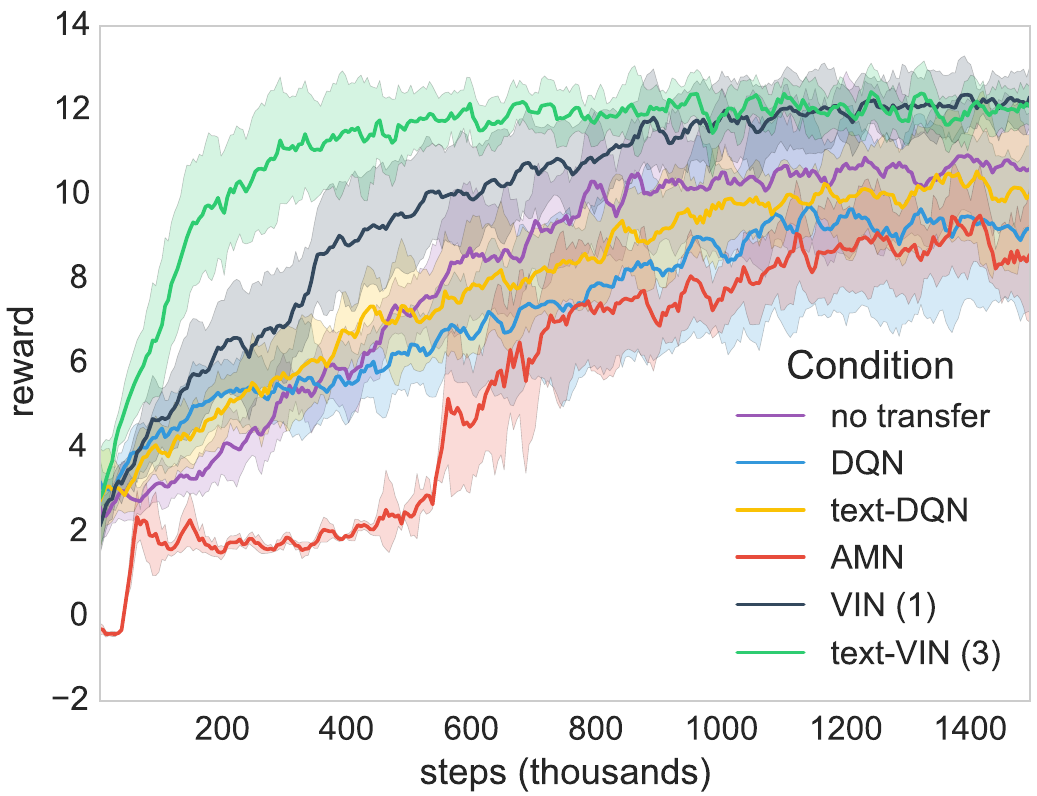}
\endminipage\hfill
\caption{Reward curves for transfer conditions (\textbf{top}) F\&E-1 $\to$ F\&E-2, (\textbf{middle}) F\&E-1 $\to$ Freeway, and (\textbf{bottom}) Bomberman $\to$ Boulderchase (best viewed in color). Numbers in parentheses for \textsc{text-vin} indicate $k$ value. All graphs were produced by averaging over 9 runs with different random seeds; shaded areas represent standard deviation.
}
	\label{fig:results}
\end{figure}

On the first metric of \emph{average reward}, \textsc{text-vin (3)} achieves a 5\% gain (absolute) over the nearest competitor AMN on F\&E-1 $\to$ F\&E-2, and a 14\% gain (absolute) over \textsc{vin (1)} on F\&E-1 $\to$ Freeway. In the case of Bomberman $\to$ Boulderchase, \textsc{text-vin} achieves an average reward of 11.17, which is 1.33 points higher than the nearest baseline of \textsc{vin}, which doesn't make use of the text. Most of this performance gap stems from the fact that our model is able to learn good policies very quickly, reusing knowledge from the source environment, and hence achieving higher rewards from the start. This fact is also evident from the sample reward curves shown in Figure~\ref{fig:results}.

\subsubsection{Initial Reward}
On the metric of \emph{initial reward}, all the transfer approaches outperform the \textsc{no transfer} baseline, except for \textsc{amn} on Bomberman $\to$ Boulderchase. \textsc{text-vin} achieves the highest numbers in all transfer settings, up to 11.5\% better than \textsc{text-dqn} on F\&E-1 $\to$ F\&E-2. This demonstrates our model's effective utilization of text descriptions to bootstrap learning in a new environment. Interestingly, \textsc{text-dqn} demonstrates good jump-start behavior on two of the conditions, but not on \text{F\&E-1} $\to$ Freeway. 

\subsubsection{Asymptotic Reward}
On the final metric of \emph{asymptotic performance}, our model improves performance over  \textsc{no transfer} and is at par or outperforms the other baselines, except on F\&E-1 $\to$ F\&E-2, where \textsc{amn} obtains a score of 1.64. This is partly due to its smoother convergence;\footnote{This fact is also noted in \cite{parisotto2016actor}} improving the stability of our model training could boost its asymptotic performance. Thus, our approach not only speeds up learning on an unseen domain, but also results in better optimal policies.

Another observation from Table~\ref{table:results} is that \textsc{text-vin} consistently outperforms \textsc{text-dqn} in all conditions. This demonstrates the importance of having a model-aware policy, which can ground text descriptions onto environment dynamics while retaining flexibility to accommodate different policies for varying task types. \textsc{text-dqn}, on the other hand, couples both knowledge of the environment and the policy into a single network, making it less suitable for transfer.

\subsubsection{Negative Transfer}
We also observe the challenging nature of policy transfer in some scenarios. For example, in Bomberman $\rightarrow$ Boulderchase, \textsc{dqn}, \textsc{text-dqn} and \textsc{amn} achieve a \emph{lower} average reward and \emph{lower}  asymptotic reward than the \textsc{no transfer} model, exhibiting negative transfer~\cite{taylor2009transfer}. Further, \textsc{text-dqn} has a lower \emph{initial reward} than a vanilla \textsc{dqn} in such cases, which further underlines the need for a model-aware approach to truly take advantage of the text descriptions for transfer.

\subsection{Multi-task Performance}
\begin{figure}[!h]
\centering
\minipage{0.6\textwidth}
\includegraphics[width=\linewidth]{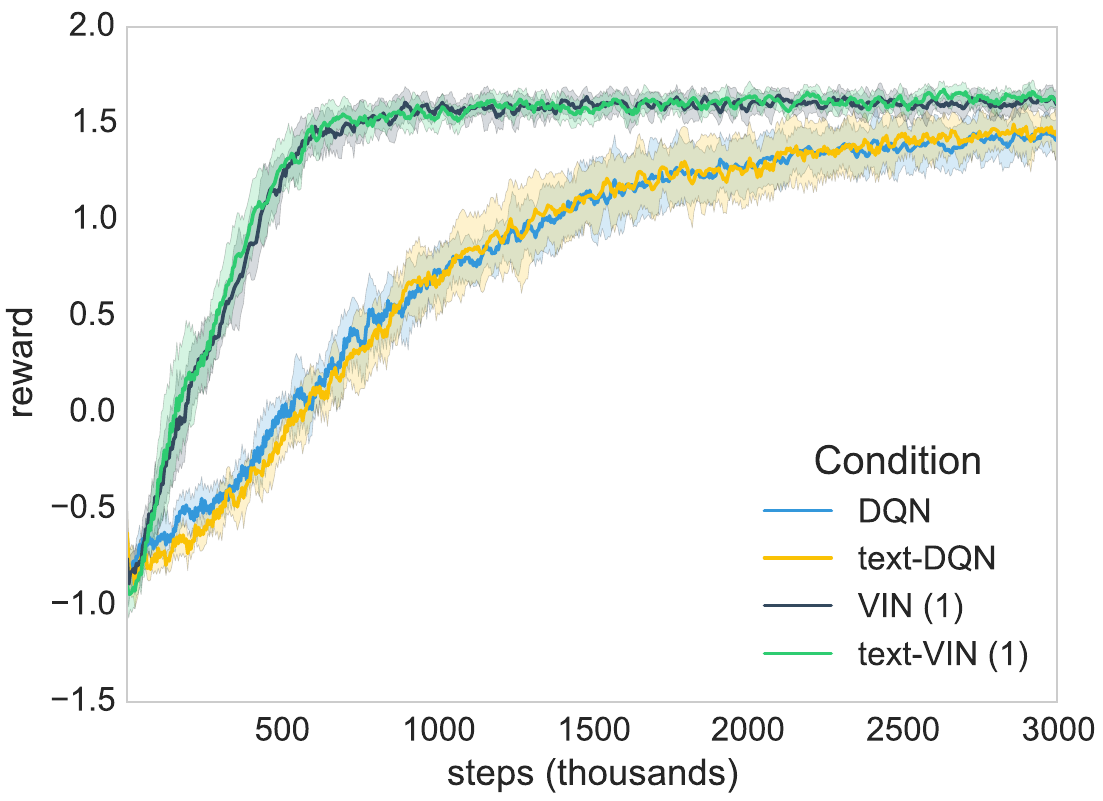}
\endminipage\hfill
\caption{Reward curve for multitask learning in F\&E-2. Numbers in parentheses for \textsc{text-vin} indicate $k$ value. All graphs averaged over 9 runs with different random seeds; shaded areas represent standard deviation.
}
	\label{fig:results-multitask}
\end{figure}

In addition to transfer, we also investigate learning in the multi-task setting~\cite{caruana1997multitask}, where the agent learns to perform multiple tasks in a single domain, simultaneously. Specifically, we train a single model, with the same set of parameters, using feedback from all the different tasks. We find that the learning benefits of our model observed in the transfer scenario also hold for multi-task learning, with benefits stemming from both the representation generator as well as the value iteration network. Table~\ref{table:multitask} details the average reward and asymptotic reward obtained by different models across twenty variants of the F\&E-2 domain. Our model is able to use the text to learn faster as well as achieve higher optimum scores, with \textsc{text-vin (1)} showing gains over \textsc{dqn} of 28.5\% and 12\% on average and asymptotic rewards, respectively. Figure~\ref{fig:results-multitask} shows the corresponding reward curves for the various models.

\begin{table}[!h]
\centering
\begin{tabular}{ c  c  c } 
\textbf{Model} & \emph{Avg.} & \emph{Asymp.}  \\ \midrule
\textsc{dqn}  & 0.80 (0.08) & 1.38 (0.07)\\
\textsc{text-dqn}  & 0.79 (0.09) & 1.45 (0.08) \\
\textsc{vin (1)} & 1.35 (0.04) & 1.61 (0.05)\\
\textsc{text-vin (1)} &\textbf{1.37} (0.03) & \textbf{1.62} (0.02) \\
\end{tabular}
\caption{Average (Avg.) and asymptotic (Asymp.) rewards for multitask learning over 20 games in F\&E-2. All numbers are averaged over 9 different runs; numbers in parentheses are standard deviations.}
\label{table:multitask}
\end{table}

\subsection{Analysis}
We further analyze the performance of our model by performing ablation studies to investigate the effects of different state and text representations, as well as a qualitative analysis of the value maps produced by the model.

\subsubsection{Effect of Factorized Representation}
We investigate the usefulness of our factorized representation by training a variant of our model using only a text-based vector representation (\emph{Text only}) for each entity, i.e. $\phi(s, Z) = v_z(s, Z)$.
We consider two different transfer scenarios -- (a) when both source and target instances are from the same domain (F\&E-1 $\rightarrow$ F\&E-1) and (b) when the source/target instances are in different domains (F\&E-1 $\rightarrow$ F\&E-2).
 In both cases, we see that our two-part representation results in faster learning and more effective transfer, obtaining 23\% higher average reward and 19\% more asymptotic reward in \text{F\&E-1 $\rightarrow$ F\&E-2} transfer (Table~\ref{table:textrep}). Our  representation is able to transfer prior knowledge through the text-based component while retaining the ability to learn new entity-specific representations quickly.

\begin{table}[!t]
\centering
\begin{tabular}{ c  c  c  c  c } 
\textbf{Condition} & \textbf{Model} & \emph{Average} & \emph{Initial} & \emph{Asymptotic}  \\ \midrule
\multirow{ 2}{*}{F\&E-1 $\rightarrow$ F\&E-1} & Text only  & 1.64 (0.02) & 0.48 (0.09) & \textbf{1.78} (0.00) \\
& Text+entity ID & \textbf{1.70} (0.02) & \textbf{1.07} (0.19) & \textbf{1.78} (0.00) \\ \cdashline{1-5}[0.5pt/5pt]
\rule{0pt}{3ex} \multirow{ 2}{*}{F\&E-1 $\rightarrow$ F\&E-2} & Text only  & 0.86 (0.01) & 0.49 (0.04) & 1.11 (0.01) \\
& Text+entity ID & \textbf{1.32} (0.06) & \textbf{0.70} (0.22) & \textbf{1.50} (0.05) \\
\end{tabular}
\caption{ Transfer results using different input representations with \textsc{text-vin (3)}. \emph{Text only} means only a text-based vector is used, i.e. $\phi(s, Z) = v_z(s, Z)$. \emph{Text+entity ID} is our full representation, $\phi(s, Z) = [v_o(s); v_z(s, Z)]$. All numbers are averaged over 9 different runs; numbers in parentheses are standard deviations.}
\label{table:textrep}
\end{table}

\begin{table}[!t]
\centering
\begin{tabular}{c  c  c  c} 
\textbf{Condition} & \textbf{Model} & \emph{BOW} & \emph{LSTM}  \\ \midrule
\multirow{ 2}{*}{F\&E-1 $\rightarrow$ F\&E-2} & \textsc{text-vin (1)}  & 1.17 (0.08) &  1.28 (0.06)\\
& \textsc{text-vin (3)} & \textbf{1.32} (0.06) &  1.20 (0.06) \\ \cdashline{1-4}[0.5pt/5pt]
\multirow{ 2}{*}{F\&E-1 $\rightarrow$ Freeway} & \textsc{text-vin (1)}  & 0.57 (0.09) & 0.63 (0.02)\\
& \textsc{text-vin (3)} & 0.57 (0.05) & \textbf{0.73} (0.01) \\ \cdashline{1-4}[0.5pt/5pt]
\multirow{ 2}{*}{Bomberman $\rightarrow$ Boulderchase} & \textsc{text-vin (1)}  & 10.09 (0.94) & 11.10 (0.22) \\
& \textsc{text-vin (3)} & 9.66 (0.77) &  \textbf{11.17} (0.44) \\

\end{tabular}
\caption{Average rewards in Bomberman $\to$ Boulderchase with different text representations: Mean bag-of-words (\emph{BOW}), or a vector generated by running an  \emph{LSTM}-based recurrent neural network over the entire sentence. All numbers are averaged over 9 different runs; numbers in parentheses are standard deviations.}
\label{table:lstm}
\end{table}

\subsubsection{Text Representation: BOW vs. LSTM}
Another question to consider is the relative impact of using LSTM vs. mean BOW to generate vector representations from the text descriptions in our model. Table~\ref{table:lstm} provides a comparison of transfer performance between these two representations on the different conditions. We observe that using an LSTM provides significantly better results on 
F\&E-1 $\rightarrow$ Freeway and Bomberman $\rightarrow$ Boulderchase, and slightly worse than BOW ($1.28$ vs $1.32$) on F\&E-1 $\rightarrow$ F\&E-2. This indicates that a good text representation which can capture linguistic compositionality works better in our model. Exploration of other recently proposed representations like the Transformer~\cite{vaswani2017attention} could lead to further improvements.

\begin{figure*}[!t]
\minipage{0.5\textwidth}
  \includegraphics[width=\linewidth]{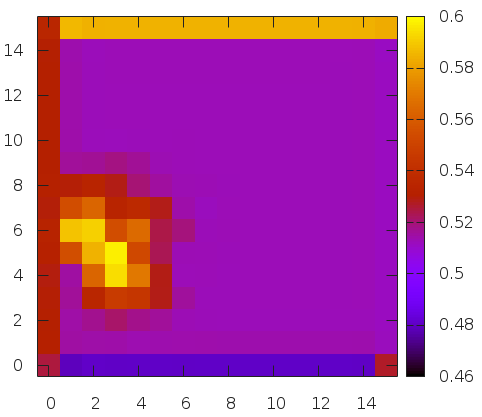}
  \caption*{(a)}
\endminipage
\minipage{0.5\textwidth}
  \includegraphics[width=\linewidth]{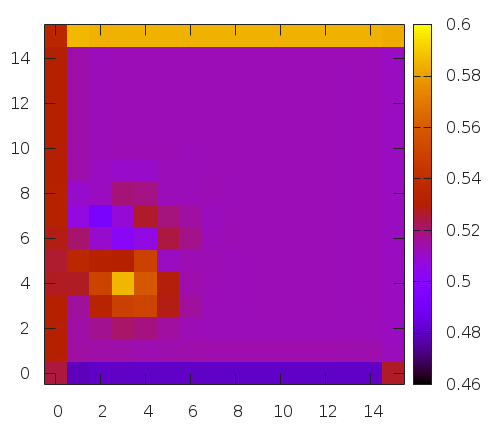}
  \caption*{(b)}
\endminipage \\
\minipage{0.5\textwidth}
  \includegraphics[width=\linewidth]{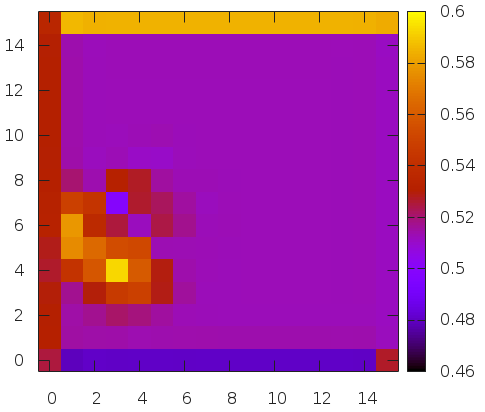}
  \caption*{(c)}
\endminipage
\minipage{0.5\textwidth}
  \includegraphics[width=\linewidth]{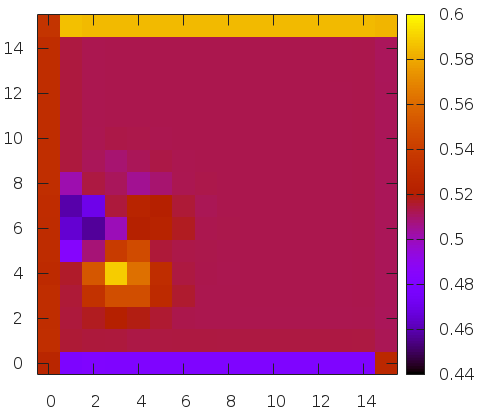}
  \caption*{(d)}
\endminipage

\caption{Value maps ($\hat{V}^{(k)}(s,Z)$) produced by the VIN module for (a) seen entity (friend), (b) unseen entity with no description, (c) unseen entity with `friendly' description, and (d) unseen entity with `enemy' description. Agent is at (4,4) and the non-player entity is at (2,6). Notice how the value of cell (2,6) changes with the type of description: higher for `friendly' and lower for `enemy' compared to the case with no description.}
	\label{fig:heatmap}
\end{figure*}

\subsubsection{Value Analysis}
Finally, we provide some qualitative evidence to demonstrate the generalization capacity of \textsc{text-vin}. Figure~\ref{fig:heatmap} shows visualizations of four value maps produced by the VIN module of a trained model, with the agent's avatar at position (4,4) and a single entity at (2,6) in each map. In the first map, the entity is known and friendly, which leads to high values in the surrounding areas, as expected. In the second map, the entity is unseen and without any descriptions; hence, the values are uninformed. The third and fourth maps, however, contain unseen entities with descriptions. In these cases, the module predicts higher or lower values around the entity depending on whether the text portrays it as a friend or enemy. Thus, even before a single interaction in a new domain, our model can utilize text to generate good value maps. This bootstraps the learning process, making it more efficient.



%% file: conclusions.tex
\section{Conclusion}
\label{sec:conclusions}
In this work, we have explored a novel method to tackle the long-standing challenge of transfer for reinforcement learning. Transferring policies is hard mainly due to the difficulty in learning effective mappings between source and target domains, often resulting in negative transfer~\cite{taylor2009transfer} as a result of incorrect mappings.
We have proposed utilizing natural language to drive transfer for reinforcement learning (RL) and shown that textual descriptions of environments provide a compact intermediate channel to facilitate effective policy transfer. In contrast to most existing systems, we have employed a model-aware RL approach that aims to capture the dynamics of the environment. For this, we utilized a value iteration network (VIN), which encapsulates the iterative computation of a value function into a single differentiable neural network. We have also introduced a two-part state representation in order to combine text with input observations.
This representation allows us to distill useful information while being robust to incomplete or noisy descriptions.

By effectively utilizing descriptions, our technique can bootstrap learning on new unseen domains. Over several empirical tests across a variety of environments, we have shown that our approach is at par or outperforms several existing systems on different metrics for transfer learning. Our model achieves up to 14\% higher average reward and up to 11.5\% higher initial reward compared to the most competitive baselines. We have also performed evaluation on a multi-task setting where learning is simultaneously carried out in multiple environments and demonstrate the superior performance of our approach.

There are several possible avenues of future work. One could explore the combination of different transfer approaches. Leveraging language for policy transfer in deep RL is complementary to other techniques such as policy reuse~\fullcite{glatt2017policy} or skill transfer~\fullcite{gupta2017learning} among other approaches~\fullcite{du2016initial,tobin2017domain,yin2017knowledge}. A combination of one of these methods with language-guided transfer could result in further improvements.
Another area for investigation is on techniques that can operate without requiring explicit one-to-one mappings between descriptions and entities in the environment. One can either learn these mappings simultaneously with the policy, or operate using descriptions that involve multiple entities and global relations.

In this work, since we factorize our input representation (using both text and direct observations), the method works at least as well as when not using the text descriptions i.e. the model could learn to rely less on the descriptions if they do not contain useful information. However, one potential case for failure could be if the text contains misleading/incorrect descriptions of the environment. Addressing this issue of robustness to adversarial inputs is another potential direction of investigation.
Finally, a major component behind our model's generalization performance is the value iteration network (VIN). However, in its current form, the VIN requires specifying a recurrence hyper-parameter $k$, whose optimal value might vary from one domain to another. Investigating models that can perform multiple levels of recurrent VI computation, possibly in a dynamic fashion, would allow an agent to simultaneously plan and act over multiple temporal scales.
